# Nature's All-in-One: Multitasking Robots Inspired by Dung Beetles

*Binggwong Leung, Stanislav Gorb, Poramate Manoonpong\**

B. Leung, P. Manoonpong

Bio-inspired Robotics and Neural Engineering Lab, School of Information Science and Technology, Vidyasirimedhi Institute of Science and Technology; Rayong, 21210, Thailand

E-mail: poramate.m@vistec.ac.th

Stanislav Gorb

Functional Morphology and Biomechanics, Zoological Institute, Kiel University; Kiel, 24118, Germany

P. Manoonpong

Embodied AI and Neurorobotics Lab, SDU Biorobotics, The Mærsk Mc-Kinney Møller Institute, The University of Southern Denmark; Odense, 5230, Denmark

**Abstract**

Dung beetles impressively coordinate their six legs simultaneously to effectively roll large dung balls. They are also capable of rolling dung balls varying in the weight on different terrains. The mechanisms underlying how their motor commands are adapted to walk and simultaneously roll balls (multitasking behavior) under different conditions remain unknown. Therefore, this study unravels the mechanisms of how dung beetles roll dung balls and adapt their leg movements to stably roll balls over different terrains for multitasking robots. We synthesize a modular neural-based loco-manipulation control inspired by and based on ethological observations of the ball-rolling behavior of dung beetles. The proposed neural-based control contains various neural modules, including a central pattern generator (CPG) module, a pattern formation network (PFN) module, and a robot orientation control (ROC) module. The integrated neural control mechanisms can successfully control a dung beetle-like robot (ALPHA) with biomechanical feet to perform adaptive robust (multitasking) loco-manipulation (walking and ball-rolling) on various terrains (flat and uneven). It can also deal with different ball weights (2.0 and 4.6 kg) and ball types (soft and rigid). The control mechanisms can serve as guiding principles for solving complex sensory-motor coordination for multitasking robots. Furthermore, this study contributes to biological research by enhancing our scientific understanding of sensory-motor coordination for complex adaptive (multitasking) loco-manipulation behavior in animals

**1. Introduction**

Although many legged robot platforms have locomotion as their primary function,[1-6] there is a growing trend toward utilizing legged robots for diverse object manipulation tasks.[7,8] These tasks involve the integration of locomotion and object manipulation, which is commonly known as a (multitasking) loco-manipulation problem. Recently, several groups of robots have been developed for object loco-manipulation. For example, one group of robots utilizes dedicated non-locomotive arms or grippers to manipulate objects.[7] These robots include the SpotMini,[9,10] Anymal,[8,11] quadruped robot,[12] LAURON V,[13] CENTAURO,[14] HyQ,[15] and HyQReal.[16] On the other hand, other type of robots use their locomotive legs to move and manipulate objects. These robots include the MELMANTIS-1 (*17, 18*),[17,18] hexapod robot (*19*),[19] quadruped robot,[20,21] PH-Robot,[22] ASTERISK,[23-25] SEA,[26] and ALPHA dung beetle-like robot.[27] In this type of manipulation strategy, the robot walks to an object and pushes it with its body (pushing), generates an impulse (kicking), performs non-prehensile lifting, or grasps and holds an object with its walking legs.[7] Although numerous approaches utilize gripper limbs to manipulate objects, employing legs to grasp and manipulate objects offers distinct benefits, particularly for manipulating objects that are larger and heavier than the robot.

Dung beetles exhibit complex and dynamic ball "rolling" behavior that is not reflected in the existing object manipulation strategies for real legged robots.[28,29] Dung beetles are capable of rolling balls of varying weights and can deal with a variety of terrains. They can transport a ball with a diameter-to-leg-length ratio of approximately 2:1[30] using an object-rolling approach to manipulate the object. This ratio is approximately twice that of existing legged robots for manipulating objects.[7,27] Consequently, this impressive form of object manipulation could provide a new way for legged robots to manipulate and transport relatively large objects with respect to their size. However, the mechanisms underlying the ability of dung beetles to reliably roll balls of varying weights and surface properties across a variety of terrains remain largely unknown and have not been fully translated to legged robots.

According to previous studies,[30,31] dung beetles roll a ball by simultaneously pushing on the ground with their front legs and walking on the ball with their middle and hind legs. For inter-leg coordination, the front legs alternately step on the ground, while the middle legs move similarly to the diagonal hind legs (Figure 1B). For intra-leg coordination, the front legs walk backward on the ground, whereas the middle and hind legs walk forward on the ball while simultaneously pushing it. When a ball is rolled on different terrains, the inter-leg coordination (gait pattern) is also adapted.[32] Based on prior research, it is hypothesized that dung beetles must adjust their motor patterns to achieve robust ball-rolling control and maintain stability when rolling balls of different types on various terrains. However, the type of internal neural control mechanisms that enable dung beetles to roll a ball robustly under a variety of conditions is still unknown. Furthermore, translation and implementation of these mechanisms to real legged robots to achieve object loco-manipulation similar to that of dung beetles has not previously been realized.

In this study, we aimed to unravel the neural mechanisms underlying how dung beetles adapt their motor patterns to steadily roll various ball types and weights over various terrain types. We synthesized a modular neural-based control with respect to four ball-rolling rules derived from our previous study on the ball-rolling behavior of the dung beetle *Scarabaeus (Khepher) Lamarcki*.[30] The synthesized neural-based control consists of various neural modules, such as central pattern generator (CPG) module, pattern formation network (PFN) module, and robot orientation control (ROC) module. An integration of the CPG and PFN modules leads to a so-called leg CPG-based control (LCPG) module that generates rhythmic motor patterns for basic locomotion, object manipulation, and their combination (loco-manipulation). Simultaneously, the ROC module controls the roll and pitch angles of the robot to achieve adaptive and robust ball-rolling control in order to deal with a variety of conditions. We implemented our modular neural-based control to a dung beetle-like robot (ALPHA)[1] and tested its loco-manipulation performance on various terrains and ball types. In addition, we investigated the use of biomechanics, including the addition of a soft material and compliant fin ray-based tarsi, [33] to improve gripping performance (leg-ball attachment) and enable passive adaptation to the terrain profile for stable ball-rolling behavior on uneven terrain. The contributions of this study are summarized as follows:

1. Adaptive (multitasking) ball-rolling behavior for large object loco-manipulation driven by modular neural-based control was derived from the ethological observations of the ball-rolling behavior of dung beetles. The control realizes two key functions. One primitive function generates rhythmic motor patterns for basic locomotion, object manipulation, and their combination (loco-manipulation). The other one focuses on balancing and adapting the robot's posture for stable ball rolling in various conditions (different ball types, weights, and terrains). The neural control solution exhibits generalizability, potentially serving as guiding principles for tackling complex sensory-motor coordination in multifunctional robots with multiple legs (like hexapod and quadruped), performing dual tasks of locomotion and large object manipulation/transportation.

---

[1]Please see Supporting Information for more details on the dung beetle-like robot.

2. The ROC module was combined with the LCPG module to stably roll a ball on various terrains (flat and uneven) and handle different ball weights (2.0 kg and 4.6 kg) and properties (soft and rigid balls). The ROC module consists of submodules including roll and pitch controls to control the roll and pitch angles of the robot while rolling a ball. The use of ROC improves the ball-rolling behavior. This allowed the robot to roll a large ball (an object size of approximately twice the robot leg length). Our robot achieved a greater object size-to-leg-length ratio than other insect-inspired robots. Our robot also achieved an overall ball-rolling speed ranging from 10 to 20 cm/s. These accomplishments represent advancements not previously demonstrated in other robots.

In the following sections, we describe a method for translating ball-rolling rules from behavioral observations on real beetles in our previous studies[30] to synthesize neural-based ball-rolling control for a dung beetle-like robot (ALPHA). The proposed bio-inspired neural-based control enables the dung beetle-like robot to successfully roll large balls across various terrains, which has not been shown by other existing approaches (see Supporting Information). We then evaluate the robot's ball-rolling performance under various conditions, such as on flat and uneven terrains and with balls of different weights and materials. Finally, we discuss the remaining issues in ball-rolling control.

## 2. Results

### 2.1. From Intra- and Inter-Leg Coordination Rules of Dung Beetle Ball-Rolling Behavior to Neural-Based Loco-Manipulation Control

In this section, we explain the intra- and inter-leg coordination mechanisms governing the ball-rolling behavior of dung beetles and describe how we transferred this knowledge to synthesize a neural-based control that generates the gait pattern for ball-rolling behavior.

According to our ethological observations,[30] dung beetles use their front legs to walk backward on the ground while simultaneously moving their middle and hind legs to push the ball. Inspired by this strategy, we propose the neural LCPG module (Figure 1D) that can generate appropriate leg movements for walking and ball-rolling. The LCPG module consists of CPG and PFN modules generating appropriate rhythmic joint movements for a single leg. In our control architecture, six distributed LCPG modules (one for each leg) are used to generate a backward walking trajectory for the front legs and a forward walking trajectory for the middle and hind legs. These trajectories are based on the intra-leg coordination in the ball-rolling behavior of dung beetles (see Materials and Methods). [30]

For inter-leg coordination, four rules from behavioral observations describe the ball-rolling gait pattern of dung beetles, and these rules result in the gait pattern illustrated in Figure 1C(i). The dung beetle alternately pushes the ground with its front legs (Rule 1), while diagonal pairs of middle and hind legs walk similarly (Rules 2 and 3). In addition, the stepping of the middle and hind legs propagates from the rear to the front (Rule 4). Two distinct support patterns are observed: 1) tripod and 2) atypical tripod support patterns. The tripod support is characterized by the simultaneous stance of legs (Left front (L1), Right middle (R2), and left hind (L3)) or (right front (R1), left middle (L2), and right hind (R3)). In contrast, the atypical tripod support is defined as when legs (L1, L2, and R3) or legs (R1, R2, and L3) are simultaneously in the stance phase (see Supporting Information). Two distinct support patterns occur because the movements of the front legs are sometimes loosely coupled with those of the middle and hind legs. As a result, we observed a similar percentage of tripod and an atypical tripod support patterns in the ball-rolling gait of dung beetles (Figure 1E).

In our preliminary experiments, we found that the tripod support pattern could allow the ball to roll more steadily than the atypical tripod support pattern (see Discussion). Therefore, we first implemented specific phase coupling between each LCPG module to allow the robot to roll the ball with a default tripod gait (see Materials and Methods). In addition, to enable ALPHA to adapt its motor patterns to stably roll a ball under different conditions, we proposed the ROC module that could modulate the robot joint movements in real-time (online intra-limb adaptation) based on the robot orientation. The ROC module uses the roll and pitch angle feedback to stabilize the middle and hind legs of the robot on the ball (Figure 1D). By doing so, it reduces the tight coupling between each LCPG, which can result in an atypical tripod support pattern.

Figure 1C(ii) shows the ball-rolling gait of the ALPHA robot (see Movie S1 in Supporting Information or https://youtu.be/ScldrZ6n5Wc), which is controlled by our newly proposed neural-based control (an integration of the LCPG and ROC modules, Figure 1D). As shown in Figure 1E, the dung beetle ball-rolling gait consists of a similar proportion of tripod and atypical tripod support patterns. In contrast, we found a higher proportion of tripod support pattern in the ball-rolling behavior of the robot. Although we encoded the tripod support pattern into the LCPG module to allow the robot to roll a ball with the default pattern, the occurrence of the atypical tripod support in the robot's ball-rolling gait could be observed. This was due to the external perturbation and adaptation of the ROC module, which could alter the robot's gait pattern to the atypical tripod support pattern (see Figure 1E, Discussion).

Further investigation on how the roll and pitch control mechanisms in the ROC module contribute to the robot's leg movement and gait pattern can be seen in Supporting Information.

## 2.2. Robot Ball-Rolling Behavior in Different Conditions under Neural-Based Loco-Manipulation Control

In this study, the capabilities of the proposed bio-inspired neural-based loco-manipulation control system of the dung beetle-like robot are evaluated under various scenarios (see Movie S1 in Supporting Information or https://youtu.be/ScldrZ6n5Wc). The robot rolled two different types of balls: a blue soft rubber fitness ball (SB) and a white rigid ball (RB) (Figure 2A). Both balls had a diameter of 60 cm (a ball size of 1.7 times the robot leg length). The soft and hard balls weighed 2.0 and 4.6 kg, respectively, while the robot weighed 4.7 kg. The rigid ball was a fiberglass sphere typically used in high-voltage wires for aircraft warnings. The robot rolled the balls on two different types of terrain (Figure 2A): 1) flat terrain and 2) uneven terrain. The flat terrain (FT) was a laminate floor, whereas the uneven terrain (UT) was a hard soil with some proportion of sand. If the robot could roll the ball over a distance of at least three meters on each terrain type, it was considered successful. Three types of biomechanical leg setups were considered (Figure 2B): 1) normal biomechanical legs covered with a textured rubber surface (NL), 2) front legs with added compliant fin-ray-based tarsi (FL, see Supporting Information), and 3) front legs with added compliant fin-ray-based tarsi and hind legs with added soft material (FL + SM). We expected that by attaching the fin-ray-based tarsi in the lateral direction of the robot, the tarsi would serve as an additional lateral support with passive adaptability to improve the ball-rolling stability. Similar to the addition of soft material to the hind legs, this modification may increase the adhesion between the leg and ball. Each biomechanical leg setup was tested with two types of neural-based loco-manipulation control (Figure 2B): 1) the LCPG module with phase synchronization to form a ball-rolling gait (LCPG), and 2) the LCPG module with phase synchronization combined with the ROC module (LCPG+ROC). Figure 2C–F show the ball-rolling speed, success rate, failure rate, and failure type of the ball-rolling tasks under different conditions.

The overall speed of the ball-rolling behavior ranged from 10 to 20 cm/s (0.3–0.6 leg-length/s). On flat terrain, the robot successfully transported the soft ball with a success rate of 100% (Figure 2D). However, the robot without the ROC module was unable to complete the task when rolling the rigid ball on flat terrain. The majority of failures occurred when the robot fell off the ball during the pitch-angle motion (Figure 2F). This occurred because the ball-rolling motion was faster than the robot motion, causing the ball to move away from the robot until the robot's hind legs slipped and it fell off the ball. The robot with the ROC module had a higher success rate than the robot without the ROC module by up to 80% (Figure 2D). When the compliant fin-ray-based tarsi were added to the front legs of the robot, the robot can roll a ball faster (Figure 2C). This suggests that the tarsi could provide more friction force between the legs and the ground to achieve stable and fast ball-rolling motion. According to the free-body diagram of the system (see Discussion), if the friction force between the front legs and ground increases ($F_{L1}$), more force will be transferred to push the ball through the middle and hind legs. As a result, the robot could roll the ball faster, however, if the ball was rolled too fast, the robot could lose control of the ball and fell off the ball. Additionally, we observed that the hind legs occasionally slipped on the ball. Therefore, soft material was added to the hind legs (Figure 2B) to increase the friction between the hind legs and the ball to prevent slipping. Owing to the increased friction between the hind legs of the robot and the ball, which increased the gripping ability and prevented the ball from rolling too fast, the robot rolled the ball at a much slower speed but was more stable (Figure 2C and 2D). On flat terrain, this resulted in a 100% success rate for rolling the rigid ball.

The robot with normal legs performed poorly on uneven terrain during the ball-rolling task (Figure 2D). The robot with the only LCPG module was unable to roll a ball over a distance of 3 m, whereas the robot with the LCPG and ROC modules could improve the success rate of by up to 20%. The remaining failure was because the uneven terrain caused the robot to tilt and fall sideways from the ball (Figure 2F). Surprisingly, attaching compliant fin-ray-based tarsi to the front legs of the robot improved the success rate by up to 80% (Figure 2D). The compliant fin-ray-based tarsi prevented the robot from tilting too far to one side and allowed the robot to maintain a stable posture while rolling the ball.

In the final condition, in which the robot rolled a rigid ball on uneven terrain (Figure 2A), the robot with normal legs was incapable of rolling the ball to the goal because the robot fell to the side (Figure 2F). Similarly, the robot with the compliant tarsi but no ROC module failed to roll the ball because it could not adapt the motor pattern to stabilize the robot's pitch angle on the ball. The robot had a 60% success rate when using the LCPG and ROC modules with compliant tarsi (Figure 2D). The addition of a soft material to the hind legs of the robot increased its success rate by up to 80%. In some trials, the robot failed to roll the ball because it pushed the ball in place (Figure 2F). This is analogous to a rigid ball rolling on flat terrain, where soft material may increase the friction between the hind legs and the ball. As a result, the robot was able to prevent the rigid ball from slipping and moving too quickly. However, it was sometimes unable to generate sufficient pushing force to overcome the uneven terrain.

Furthermore, we also analyze the robot's orientation distribution in roll and pitch angles as the robot operates under various conditions to assess the limitations of the proposed control system (see Supporting Information).

## 2.3. Emergent Behavior

In addition to the robot's ability to roll a ball under various conditions, we discovered an interesting behavior when it rolled a ball toward a wall (Figure 6A and 6B). The entire robot rotated slowly in the yaw direction until it became parallel to the wall. This emergent behavior is a result of the ball's interaction with the wall and the adaptability of our neural control system. Figure 6C clearly shows that activation of the roll and pitch controls adaptively increased when the ball came into contact with the wall to maintain a stable ball-rolling behavior. As the ball struck the wall, the robot's gait pattern was perturbed, and the control system could adapt to handle it (Figure 6D and 6E). In addition, we compared the activation percentages of the roll and pitch controls when the ball collided with a wall versus the ball rolling on flat and uneven terrain. The roll and pitch controls were more frequently activated in the wall condition than on the flat and uneven terrain (Figure 6F and 6G). This could imply that the wall disturbs the system more than the uneven and flat terrain. In the experiments, both soft and rigid balls were evaluated (Figure 6H). The robot had a success rate of 80% when rolling the soft ball and 60% when rolling the rigid ball. The high success rate of rolling the soft ball could be that the soft ball was more compliant with the wall impact than the rigid ball.

## 3. Discussion

## 3.1. Robot Ball-Rolling Behavior

In a preliminary experiment, we discovered that a tripod support pattern allows the ball to roll more steadily than an atypical tripod support pattern. This is because the moment produced by the middle and hind legs causes the ball to rotate in a particular direction. This requires the front legs to establish a tripod support pattern to maintain the robot's stability. The rolling behavior of the ball is illustrated in Figure 3 when the legs R3 and L2 are in contact with the ball. The robot's weight ($mg$) is distributed to the legs R3 and L2 via the forces $F'_{R3}$ and $F'_{L2}$. The amplitudes of the vertical forces acting on the ball, $F'_{zR3}$ and $F'_{zL2}$, differ because of the different leg placement positions on the ball. The ball provides direct support to the hind leg because that leg is positioned on the upper portion of the ball, whereas the middle leg is positioned on the side of the ball. The ball supports the weight of the robot with its hind leg structure, allowing the hind leg to generate a greater vertical force. In contrast, the middle leg can only generate force on the ball through the leg contact friction, which is much smaller. Consequently, the sum of the moments $F'_{zR3}l_1$ and $F'_{zL2}l_2$ causes the ball to rotate in the direction of the hind leg placement. Therefore, the front leg on the same side as the hind leg should be used to support the body and maintain the stability of the robot. This is the reason for using a tripod gait as a default ball-rolling gait. Consequently, we implemented a specific phase coupling between each LCPG module to allow the robot to roll a ball with a tripod gait. Although we designed the neural control to allow the robot to roll a ball with a tripod gait, we still observed an atypical tripod support in the robot's ball-rolling gait. This atypical tripod support was due to the external perturbation and adaptation from the ROC module, which could alter the gait pattern into an atypical tripod support pattern (Figure 1E). In other words, the ROC module introduces a control mechanism and adaptability to reduce the tight phase coupling strength between the LCPG modules.

We found that our robot and real dung beetles could achieve almost equivalent object-size-to-leg-length ratios; the species of dung beetle we observed achieved a ratio of 2 and our robot achieved a ratio of 1.71(≈2). Compared with other observations,[34] we noticed that *Circellium bacchu* dung beetles could roll larger balls with larger object-size-to-leg-length ratio.

Furthermore, based on our observations (Figure 4), the dung beetle could roll a ball 10 times its own weight using both the tripod and atypical tripod gaits (Figure 1E). In other words, the dung beetles have an object-to-body weight ratio of 10:1 (Figure 4), whereas our robot has a ratio of approximately 1:1. We hypothesize that the ball-rolling gait could be influenced by the object-to-body weight ratio. Because of the greater inertia of a heavier ball, it is more difficult to tilt it; consequently, a dung beetle can utilize either a conventional tripod or an atypical tripod to achieve stable rolling behavior. Additionally, the scaling laws may also effect the object-to-body weight ratio that the dung beetle and robot can achieve. In the animal kingdom, smaller animals (like ants) are relatively "stronger" because the body mass-to-force is scaling proportionally according to the power of 3/2. As a result, the scaling laws may limit the object-to-body weight ratio (Figure 4), that larger beetles/robots can reach.

In addition, the difference in contact mechanics may influence the stability of the ball-rolling behavior. Dung beetles create a mechanical interlocking contact mechanism by inserting their stiff sharp pointed tibial spurs and pretarsal claws middle and hind legs into a soft dung ball. This is unlike the contact between the robot's legs and the ball, which is solely the result of contact-area-mediated friction (adhesion-mediated friction).[35,36] Theoretically, surface friction should have a lower friction coefficient than mechanical interlocking. By having robust contact mechanics on both the middle and hind legs like the dung beetle, the body weight can be distributed more evenly on the ball, making the ball rolled by dung beetles more resistant to tilting than those in our robot experiments.

## 3.2. Legged Robots for Object Loco-Manipulation

The literature describes various insect-like robots that can perform object transportation tasks. When comparing this group of robots, having similar morphologies to that of our robot, it becomes obvious that our methodology, for the first time, successfully achieves a remarkable object size to robot leg length ratio of approximately 1.71 (see Figure 4A and 4B). The hexapod robot with the closest performance (object size to leg length ratio of 1.44) manipulates an object by pushing or grasping it with two legs and walking with the remaining legs. However, it has only been tested on flat terrain (see Figure 4). The other robots, which have achieved object size to robot leg length ratios of less than 1.0, mostly manipulate objects using the double-leg grasping and walking method. For example, the dung beetle-like robot (ALPHA) in the previous study can achieve an object size to robot leg length ratio of approximately 0.7. Our proposed approach for transporting large objects over uneven terrain can be attributed to its use of the ball-rolling approach.

Furthermore, our bio-inspired design demonstrated superior task efficiency (considering all three criteria: payload capacity (measured by weight ratio), payload size (measured by size ratio), and object transportation speed (via Froude number)), particularly on uneven terrain, compared to other robot types (see Figure 4A and 4B). The unique feature of our design allows the robot to attach or morph most of its parts (four legs) around large (spherical) objects, while using the remaining two legs solely for locomotion—similar to the behavior of dung beetles (Figure 4C). This approach also offers significant advantages in terms of space efficiency (see Supporting Information), especially for object transportation in limited space environments (Figure 4C). In contrast, other robot designs, such as those with hexapod or quadruped configurations, typically require more space for object transportation. Typical hexapod robots often extend two legs to grasp or push the object, while quadruped robots usually place the object on their back (Figure 4C). Transporting large, spherical objects like the one shown here can be particularly challenging using traditional strategies due to the object shape and weight. By employing the robot's four legs as grippers and maintaining a close proximity to the grasped object, our strategy can prevent the object from slipping or falling away from the robot. At the same time, the object serves as a third support point, providing stability to the robot while performing object transportation. Additionally, this strategy minimizes the robot's environmental footprint during transportation. This means that the robot occupies less physical space, causes minimal disruption to its surroundings, and can operate more efficiently in confined areas while transporting objects

While ideal spheres with smooth surfaces roll easily due to their single point of contact with the ground, which minimizes friction during rolling motion, other complex objects—such as balls with convex points (uneven surfaces), ellipses, or boxes—have larger or more irregular contact areas, increasing friction and making rolling more difficult. Additionally, a sphere remains naturally stable while rolling because its center of mass is always directly above the point of contact, allowing it to maintain steady motion. In contrast, these other complex objects tend to be less stable, as their center of mass may not always align with the point of contact, making them more prone to tipping or wobbling. As a result, greater rolling force and object motion prediction models are required. These models predict the irregular motions of objects based on physical robot-object interactions.[37] This information can be used to adapt the robot's leg movements, ensuring stability. Alternative object transportation strategies can be considered to handle complex objects. For example, using hind legs to grasp and push the object (as described in [27]) or carrying the object on the robot's back could be more effective than rolling if the object's size is manageable (see Figure 4C). Therefore, to efficiently control the robot in transporting various complex objects, multiple transportation strategies (such as rolling, pushing/grasping, or carrying, as shown in Figure 4C) should be implemented. Visual object recognition can then be applied to identify the target object's shape/size and activate the appropriate transportation mode.

This work contributes to the field of robotics by offering insights into key control functions essential for sensory-motor coordination in complex, adaptive dual tasks involving locomotion and large object manipulation/transportation. One primitive function generates rhythmic motor patterns for basic locomotion, object manipulation, and their combined operation (loco-manipulation). The other, an additional module, focuses on balancing and adapting the robot's posture for stable ball rolling in various conditions (different ball types, weights, and terrains). This study proposes a solution to implement these functions through two main neural-based control mechanisms (LCPG and ROC) with a distributed control architecture. The LCPG mechanism leverages CPG and PFN submodules, while the ROC mechanism integrates roll and pitch control submodules. Our neural control solution suggests a possible mechanism for the complex sensory-motor coordination underlying biological ball-rolling behavior. In other words, this study contributes to a refined scientific understanding of adaptive loco-manipulation behaviors in animals.

Our method to complex ball manipulation for legged robots differs from recent machine learning-based methods [38,39] in several key aspects. First, our method enables the robot to simultaneously manage with multiple environment interactions (i.e., ground-ball-robot interactions). Specifically, the robot dynamically interacts with a moving ball (a dynamic environment) using its middle and hind legs, while engaging the ground (a static environment) using its front legs, whereas previous studies [38,39] mainly addressed single environment interactions (i.e., ball-robot or ground-robot interactions) where all legs interacting with the ball or the ground.

Second, our method can handle a wider range of ball types (soft and rigid) and terrain conditions (even and uneven), whereas prior works focused on specific scenarios, such as walking atop a soft ball on flat ground[38] or rotating a soft ball in a circus-like manner while the robot remained statically lying on the ground.[39] Our control strategy, inspired by dung beetles, is particularly well-suited for transporting heavy objects, as it uses the robot's leaning body to generate greater force transfer (Figure 1), enabling it to overcome uneven terrain more effectively. In contrast, existing control strategies, such as walking on the ball, [38] may struggle to apply sufficient force to propel the ball forward, especially on uneven terrain.

Lastly, unlike machine learning-based methods that typically require extensive feedback (e.g., at least 24-dimensional joint position and velocity feedback) and retraining for different robots, our method operates with minimal sensory input—requiring only 2-dimensional body orientation feedback without joint data. Its analyzable and understandable control mechanisms offer generalizability for multifunctional robots performing dual tasks.

For instance, our control method can be applied to object rolling in a quadruped robot (see Supporting Information and Movie S2 or https://www.youtube.com/watch?v=tB6kJE8yfQg).

Recent approach to solve loco-manipulation task has increased the payload of quadrupedal robots by adopting the prismatic quasi-direct-drive (QDD) leg. [40] This strategy allows the robot to withstand a 50 kg payload while walking, resulting in an object-to-robot weight ratio (≈1) comparable to the ball rolling approach in our study (Figure 4). In the future, it would be interesting to apply the quasi-direct-drive (QDD) mechanism to our robot system for increasing an object-to-robot weight ratio in dynamic ball rolling. The dynamic ball-rolling approach of robots can offer practical applications across various fields, including construction, search and rescue, inspection, and agriculture. For instance, robots can transport hollow plastic spherical balls used in Bubble Deck slab systems for building construction. [41] By placing objects inside a hollow ball, serving as a container and a protective shell, this enables the robot to deliver food and medicine to survivors in search and rescue scenarios or transport inspection tools for inspectors in limited space environments. The principle of robot rolling objects could also be extended to rolling round hay bales in agricultural fields. [42]

## 4. Materials and Methods
### 4.1. Leg CPG-Based Control (LCPG) Module

The LCPG module is used to control the ball-rolling posture and rhythmic movement for the ball-rolling behavior (Figure 5). The LCPG module was created using several neural modules based on our previous study,[27] including the central pattern generator (CPG) and pattern formation network (PFN) modules. To produce a tripod gait for inter-leg coordination in ball-rolling behavior, we activate the CPG module of each leg at distinct times. The CPG module of legs R1, L2, and R3 is activated first, while the CPG module of legs L1, R2, and L3 is delayed by a certain number of time steps (i.e., $\tau = 70$ time steps). Consequently, the output signals between these two sets of legs will have an approximately $\pi$-radian phase difference (Figure 5C), resulting in a tripod gait pattern for the robot movement.

A PFN module is used to define the intra-leg coordination of each joint in a leg. In this study, the PFN first modifies the CPG output signals to produce an asymmetrical pattern of short-swing and long-stance leg movements at the $PC_1$ and $PC_2$ neurons. These neurons function as CPG post-processing neurons. The remaining part of the network ($P_1,...,P_{12}$), inspired by the phase switching network (PSN) module in a previous study,[27] is used to generate forward or backward leg movement trajectories. By setting the value of an input neuron (i) to 1, the PFN module outputs signals that cause the front legs to walk backward on the ground. The input neuron of the PFN module is set to zero to generate a forward walking motion for the middle and hind legs.

Finally, the PFN projects its output signals to the BC, CF, and FT motor neurons (Figure 5A and 5B). The amplitude of the signals is determined by the weight connections between the PFN module and the motor neurons ($W_{BC}$, $W_{CF}$, $W_{FT}$). The biases of the motor neurons ($b_{BC}$, $b_{CF}$, $b_{FT}$) are used to define the fixed position for the joint command, which results in the default posture of the robot. All of the weight connections and bias values are provided in Supporting Information.

All neurons are modeled as discrete-time non-spiking neurons, connected by synapses, and updated at a frequency of 60 Hz. The activity of each neuron develops according to the following:

$$a_i(t+1) = \sum_{j=1}^{n} w_{ij} o_j(t) + b_i; \quad i = 1, \ldots n, \tag{1}$$

where $n$ denotes the number of units, $b_i$ is an internal bias term or stationary input to neuron $i$, and $w_{ij}$ is the strength of the synaptic connections from neurons $j$ to $i$. The output, $o_i$, of all neurons uses a hyperbolic tangent (tanh) transfer function ($o_i = tanh(a_i)$, [-1, 1]), except for the $PC_1$ and $PC_2$ neurons, which employ a step function. The motor neurons employ piecewise linear transfer functions. The synaptic weights are empirically adjusted to achieve the ball-rolling behavior.

**4.2. Robot Orientation Control (ROC) Module**

In this study, we propose an ROC module for balancing and stabilizing the robot during ball-rolling (Figure 5). It consists of two sub-modules that control two constraints essential for accomplishing the ball-rolling behavior: 1) the robot's roll angle and 2) the robot's pitch angle (Figure 5D). The roll and pitch controls use feedback from an IMU sensor (Model: 1044-Phidget Spatial Precision 3/3/3) and output signals to modulate the motor neurons in each leg.

*4.2.1. Roll Control*

Roll control is employed to prevent the robot from leaning too far to the left or right. According to the roll control concept, if the robot tips too far to one side, one of the front legs needs to push against the ground to rebalance and support the robot to return to the stable rolling posture (Figure 5G). Figure 5F show the robot roll control diagram. The IMU sensor provides the control loop with the roll angle feedback ($\phi_f(t)$) (Figure 5D). The roll error signal ($e_\phi(t)$), which is produced by comparing this value to the reference roll angle ($\phi_r$), is calculated as follows:

$$e_\phi(t) = \phi_r - \phi_f(t) \tag{2}$$

The roll error signal ($e_\phi(t)$) is then projected through two pathways to either modulate the movement of the front legs or inhibit the movement of the middle and hind legs (Figure 5E). The roll error signal for the front legs is processed using ReLU functions and then sent to the motor neurons, as shown in the following equations:

$$m_L(t) = ReLu(-e_\phi(t) - b), \tag{3}$$
$$m_R(t) = ReLu(e_\phi(t) - b), \tag{4}$$
$$CF_0(t) = CF_0(t) + k_{CF} m_L(t), FT_0(t) = FT_0(t) + k_{FT} m_L(t), \tag{5}$$
$$CF_3(t) = CF_3(t) + k_{CF} m_R(t), FT_3(t) = FT_3(t) + k_{FT} m_R(t), \tag{6}$$

where $m_L(t)$ and $m_R(t)$ represent the left and right modulations of the CF and FT joints of the left and right legs, respectively; $b = 10$ and represents the bias for the ReLU activation neuron. As a result, the roll control will modulate the movement of the front legs when the roll angle exceeds 10°. $ReLU()$ represents the ReLU function. $k_{CF}$ is a gain for CF joints, whereas $k_{FT}$ is a gain for FT joints. $CF_0(t)$ and $FT_0(t)$ are the CF and FT joints of the left front leg (L1) and $CF_3(t)$ and $FT_3(t)$ are the CF and FT joints of the right front leg (R1), respectively.

If the robot tilts by more than 10°, the roll control outputs a left or right modulation ($m_L(t)$, $m_R(t)$) to drive the CF and FT motor neurons of the left front legs (as shown in Equation (5)) or right front legs (Equation (6)), respectively. The output signal from this pathway will stretch the leg to push against the ground, which will cause the robot to tilt in the opposite direction to regain stability (Figure 5G).

In contrast, in another pathway, the error signal is sent to the absolute function, which is activated if the value is greater than 10 (Figure 5F). Therefore, the shunting inhibition mechanism will inhibit the movement of the middle and hind legs if the roll angle of the robot exceeds 10°. Consequently, the inhibited joint will be fixed at a particular position based on the bias of the motor neuron.

The combination of these two pathways allows the robot to prevent the middle and hind legs from pushing the ball while it is unstable (tilting sideways). When the robot is tilted, its front legs will extend to the ground to maintain its stability on the ball. After the robot stabilizes, its middle and hind legs will begin to move and push the ball again.

*4.2.2. Pitch Control*

The pitch control of the robot is intended to maintain the pitch orientation and prevent the robot from tipping over or falling off the ball (Figure 5H). Figure 5I shows the concept of our proposed solution. If the pitch angle feedback matches the reference pitch angle, then all legs are moved with the normal amplitude. However, if the robot starts climbing up the ball, the amplitude of the front leg will be reduced to prevent it from tipping over. However, if the robot begins to fall off the ball, the amplitudes of the middle and hind legs will be reduced to prevent the robot from falling off the ball.

This study proposes a solution by modulating the amplitude of the leg trajectories by shunting the motor neuron activities (Figure 5E). By changing the leg movement amplitude according to the pitch angle, the moment of force acting on the robot can be changed to control the robot's pitch angle. The following equations describe the pitch angle control of the robot.

$$e_\theta(t) = \theta_r - \theta_f(t), \tag{7}$$

$$sf_F(t) = \alpha \cdot \min(\frac{e_\theta(t)}{\delta} + 1, 1) + (1 - \alpha) \cdot sf_F(t-1), \tag{8}$$

$$sf_B(t) = \alpha \cdot \min(\frac{-e_\theta(t)}{\delta} + 1, 1) + (1 - \alpha) \cdot sf_B(t-1), \tag{9}$$

$$BC_{0,3}(t) = BC_{0,3}(t) \cdot sf_F(t), \tag{10}$$

$$BC_{1,2,4,5}(t) = BC_{1,2,4,5}(t) \cdot sf_B(t) \tag{11}$$

where $\theta_r$ is the reference pitch angle, which is the angle at which the robot stands statically in its initial rolling posture (Figure 2A); $\theta_f(t)$ is the pitch angle feedback from the IMU sensor.

The pitch angle error ($e_\theta(t)$) is calculated using Equation (7). The pitch angle error is then processed through a linear piecewise activation function neuron and output as the front and back shunting gains (*sf<sub>F</sub>(t), sf<sub>B</sub>(t)*) (Equation (8) and (9)). The shunting gains are limited to the range of [0, 1]. It is also low-passed by the recurrent connection (Figure 5H). $\alpha$ is a recurrent weight for shunting gains which functions similarly as a low-pass filter. $\delta$ represents a constant value for tuning the slope of the linear activation neuron. In our experiments, $\delta$ is set to 10, meaning that if the pitch angle feedback error is equal to +10, the front shunting gain will became zero. If the pitch angle feedback error is equal to -10, the back shunting gain will became zero. $BC_{0,1,2,3,4,5}(t)$ represent the BC joint of legs L1, L2, L3, R1, R2, and R3, respectively. The front shunting gain, $sf_F(t)$, is used to reduce the BC joint movement of the front legs, whereas $sf_B(t)$ is used to reduce the joint movement of the middle and hind legs (Equation (10) and (11)).

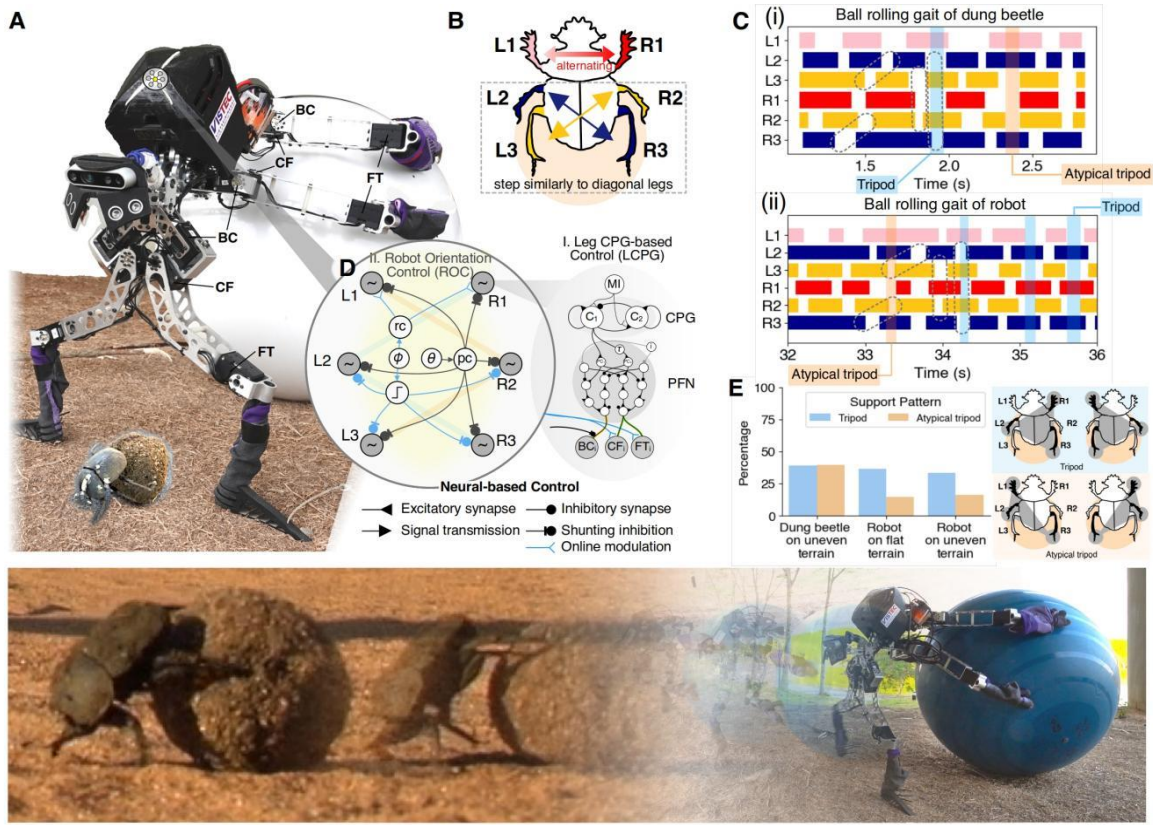

**Figure 1. Translating intra- and inter-leg coordination rules of dung beetles to neural-based ball rolling control.** (**A**) Dung beetle-like robot (ALPHA) and its biological counterpart during ball-rolling behavior.[30] The dung beetle snapshots are from a video courtesy of Marie Dacke and Emily Baird. The biomechanical structure of ALPHA is based on South African ball-rolling dung beetles.[43] It has six legs each with three joints (BC: Body-Coxa, CF: Coxa-Femur, FT: Femur-Tibia). (**B**) Ball-rolling rules derived from the behavioral investigation of real dung beetles. [30] Red arrow indicates that the front legs alternately stepping on the ground. Yellow and blue arrows indicate that a pair of legs with the same color tends to swing and stand at the same time. (**C**(i)) and (**C**(ii)) Ball-rolling gaits of a dung beetle and ALPHA on uneven terrain, respectively. (**D**) Neural-based control inspired by the ball-rolling rules. Bio-inspired neural-based control mechanisms also suggest a possible option for the sensory–motor coordination underlying the ball-rolling behavior of dung beetles. (**E**) Percentages of tripod and atypical tripod support patterns found in the ball-rolling behavior of dung beetles and ALPHA using the proposed neural-based control (see Movie S1 in Supporting Information or https://youtu.be/ScldrZ6n5Wc).

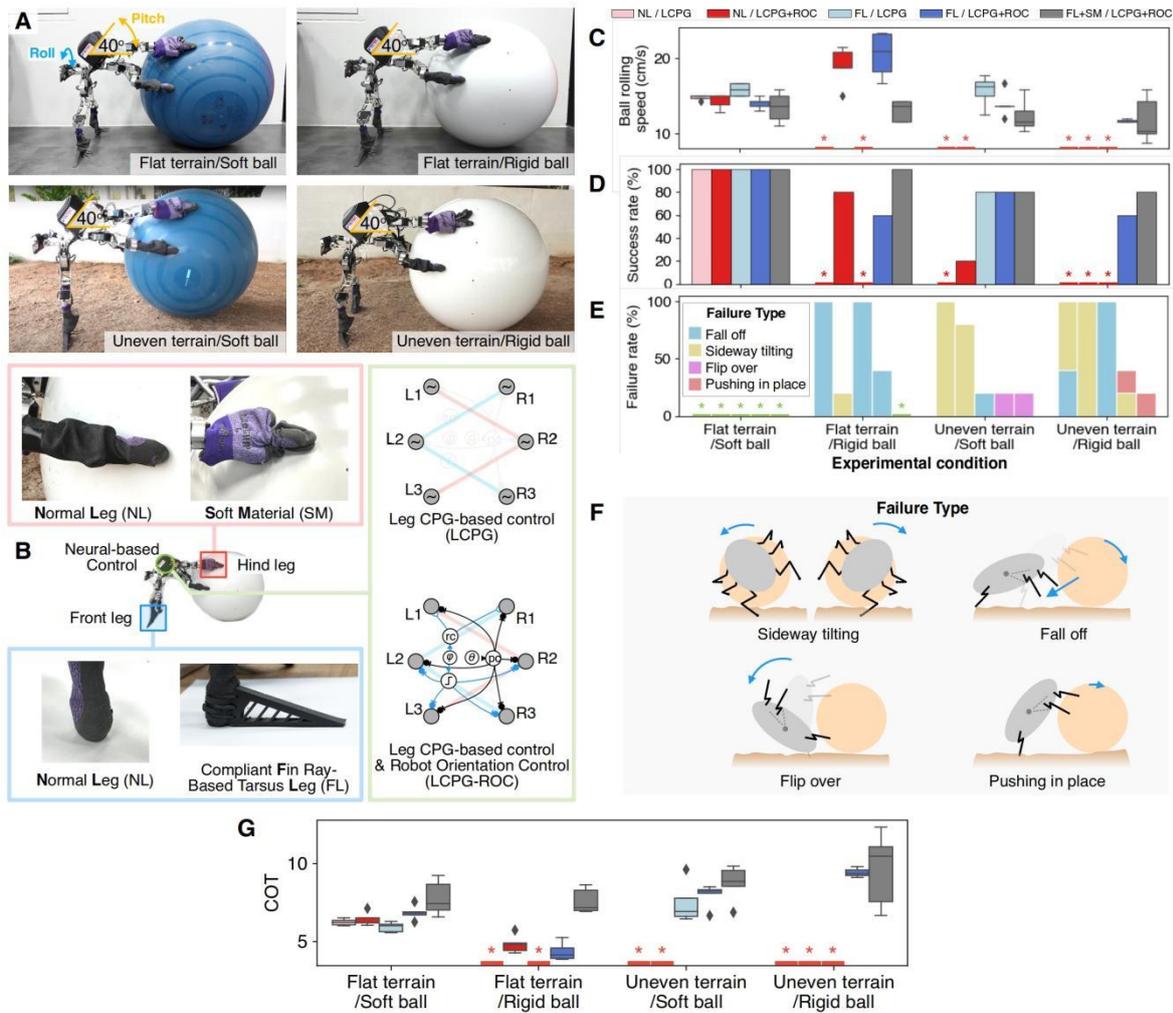

**Figure 2. Performance of the dung beetle-like robot (ALPHA) in ball-rolling loco-manipulation under various conditions.** (**A**) Illustration of rolling a 2 kg soft rubber ball and a 4.6 kg fiberglass ball on flat and uneven terrains. (**B**) Types of neural-based control and biomechanics of ALPHA in the ball-rolling experiments. (**C**) Ball-rolling speed. NL/LCPG means the robot with normal legs using the LCPG module. NL/LCPG+ROC means the robot with normal legs using the LCPG and ROC modules. FL/LCPG means the robot with compliant fin ray-based tarsi attached at the front legs using the LCPG module. FL/LCPG+ROC means the robot with compliant fin ray-based tarsi attached at the front legs using the LCPG and ROC modules. FL+SM/LCPG+ROC means the robot with compliant fin ray-based tarsi attached at the front legs with soft material at the hind legs using the LCPG and ROC modules. (**D**) Success rate for rolling a ball at a distance of at least 3 m. Red stars indicate that the robot was unsuccessful in ball rolling. (**E**) Failure rate under different conditions in the ball-rolling experiments. (**F**) Failure types observed in the ball-rolling experiments. Green stars represent conditions with 100% success rates. Each condition was

evaluated five times. (**G**) Cost of Transport (COT) of ball-rolling behavior in different conditions.

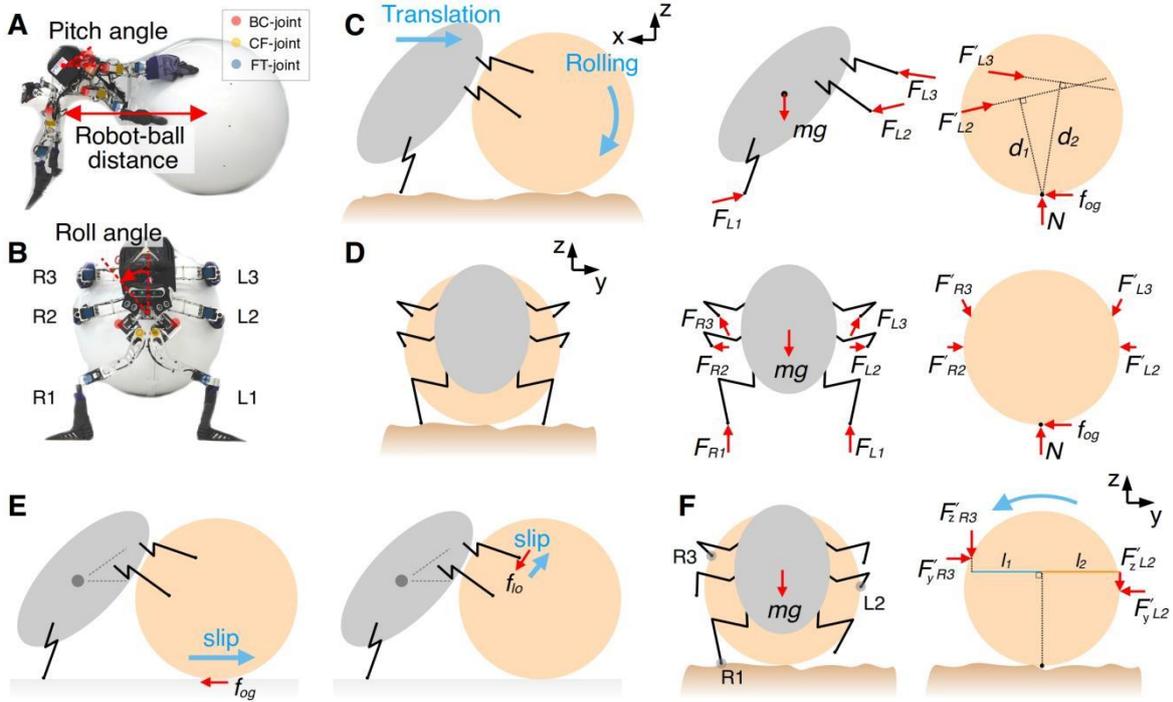

**Figure 3. Free body diagram of the robot ball-rolling system.** (**A**) Side view of the rigid ball-rolling behavior. (**B**) Front view of the rigid ball-rolling behavior. (**C**) Free body diagram of the side view. (**D**) Free body diagram of the front view. (**E**) Illustration of the slip condition in the rigid ball-rolling behavior. (**F**) Illustration of the contact force between the middle and hind legs and the ball during the rolling behavior. L1, L2, L3, R1, R2, and R3 represent the left and right front, middle, and hind legs of the robot, respectively. $F_{L1}, F_{L2},...,F_{R3}$ represent the forces from the ground or ball acting on the robot legs. $F'_{L1}, F'_{L2},...,F'_{R3}$ represent the forces from the legs acting on the ball. mg is the weight of the robot. $N$ is the normal force from the ground acting on the ball. $f_{og}$ is the object–ground friction. $f_{lg}$ is the leg–ground friction.

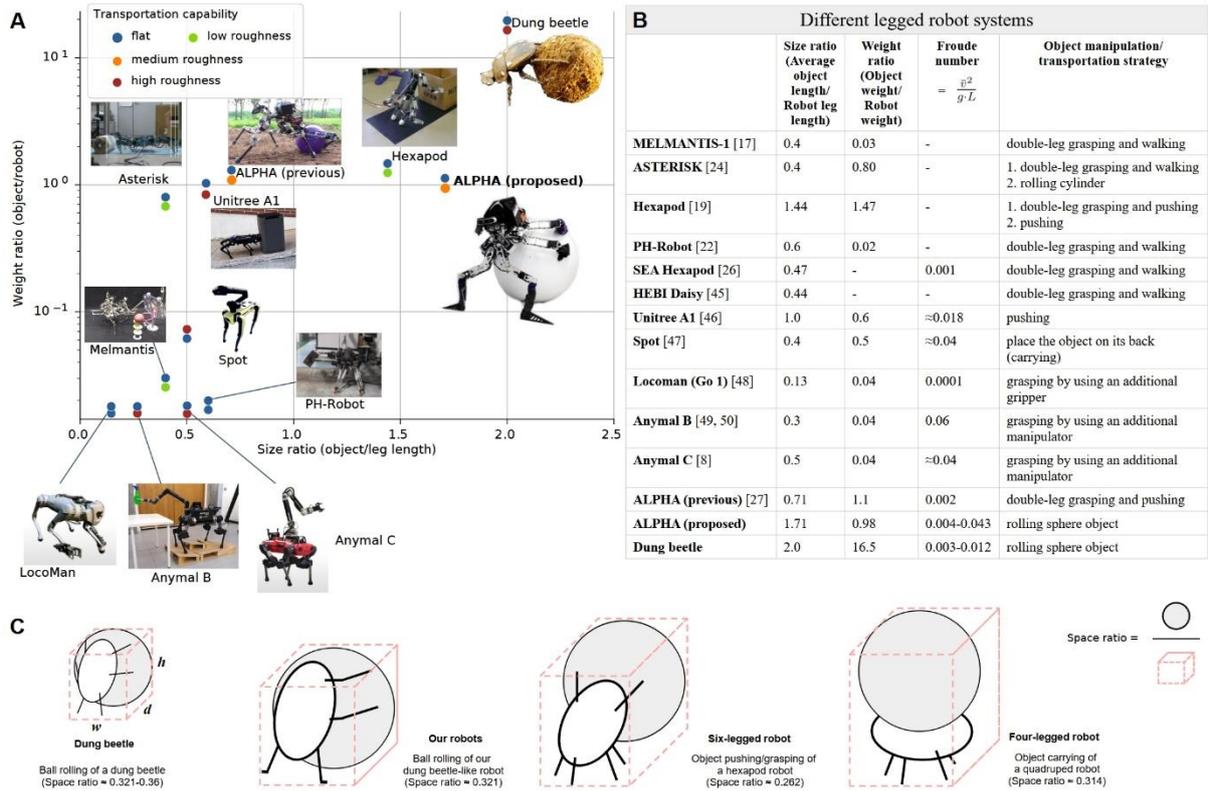

**Figure 4. Comparison of the object-transportation capability of different robot systems.** (**A**) The chart compares dimensionless parameters including i) the size ratio (object size to dung beetle or robot leg length), ii) weight ratio (object weight to dung beetle or robot weight), and iii) terrain roughness ratio (terrain height or slope difference to dung beetle or robot leg length, see Supporting Information for more detail of the terrain roughness calculation and level). The images of walking robots are reproduced from [8, 17, 19, 22, 24, 27, 46, 47, 48, 49] with permission. (**B**) Object-transportation capability of different legged robot systems. The table also shows object-transportation capabilities of dung beetles and the dung beetle-like robot (ALPHA). The observed dung beetle species is *Scarabaeus (Khepher) Lamarcki*. The Froude number[44] is calculated based on the average ball-rolling speed ($\bar{v}$, unit: *m/s*), gravitational constant (*g*, value: 9.81 *m/s²*), and robot leg length (*L*, unit: *m*). The average object length is $\bar{L}_{obj}$. The average object length is calculated as an average of the width, length, and height of the object. (**C**) Different possible large object transportation strategies using existing robot body parts (excluding additional manipulator/gripper installation): 1) rolling-based object transportation, 2) pushing/grasping-based object transportation, and 3) carrying-based object transportation. The number associated with each strategy represents its object-to-space ratio, calculated as the ratio of the transported object's volume to the total occupied space (including both the object and the robot). Here, a large object size (i.e., a size ratio of 1.5) is considered for the space ratio calculation (see Supporting Information for more detail of the calculation). A higher ratio indicates greater space efficiency. In principle, the rolling-based object transportation is particularly efficient for moving heavy or large objects over long distances. The pushing/grasping strategy can be more suitable for smaller objects. The carrying-based object transportation is ideal for smaller, lighter objects or when precision is required.

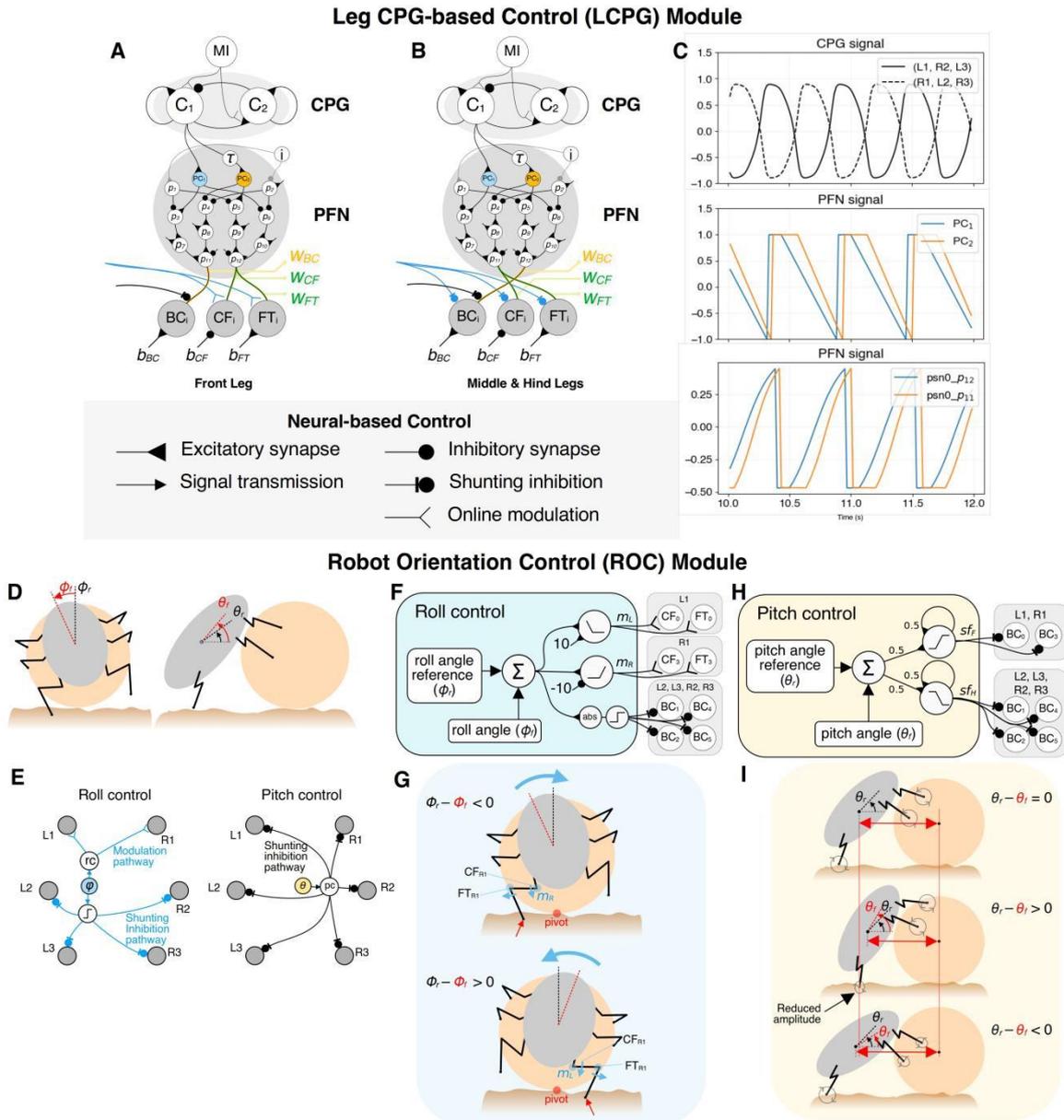

**Figure 5. Neural loco-manipulation control of a dung beetle-like robot. The control consists of LCPG and ROC modules.** (A) Control architecture of the front leg. Note that the left and right front legs have the same control architecture. (B) Control architecture of the left and right middle and hind legs. (C) Output signals from the CPG submodule in the LCPG module of each leg, inputs of the PFN module of leg L1 (neurons PC1 and PC2), and outputs of the PFN module of leg L1 (neurons p11 and p12). (D) Ball-rolling behavior and parameters. (E) ROC module consisting of roll and pitch control. (F) Roll control. (G) Illustration of the roll control concept. (H) Pitch control. (I) Illustration of the pitch control concept.

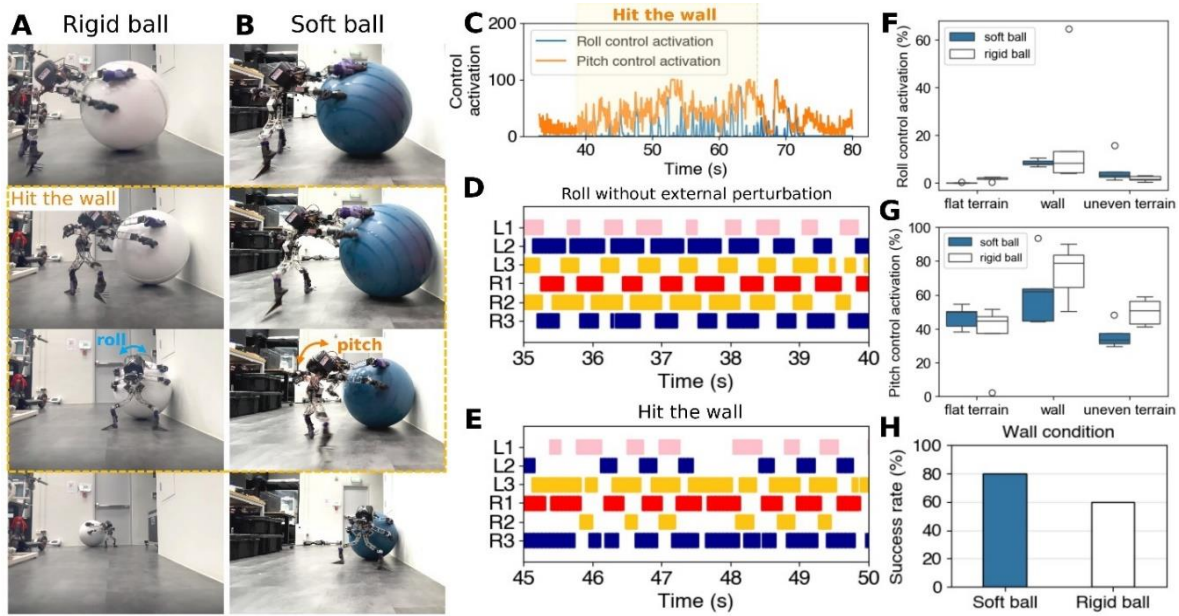

**Figure 6. Wall interaction and adaptation of the ball-rolling behavior.** (**A**) Rigid ball rolling. (**B**) Soft ball rolling. (**C**) Roll and pitch control activation percentages of the soft ball rolling. (**D**) Gait pattern while the robot rolled the soft ball without external perturbation. (**E**) Gait pattern when the soft ball contacted and moved along the wall. (**F**) and (**G**) Roll and pitch control activation percentages during ball-rolling behavior under flat, uneven, and wall conditions. Note that the variability in roll control activation was mainly influenced by the significant disturbances caused by wall interactions, which could lead to pronounced side-to-side tilting. As a result, we observed a wide range of roll control activations across the wall condition trials, with one trial exhibiting more large roll instability than the others. (**H**) Success rates for rolling the soft and rigid balls in the wall condition.

**Supporting Information**

Supporting Information is available from the Wiley Online Library or from the author.

**Acknowledgements**

This research was supported by the Startup Grant-IST Flagship research of VISTEC on Bio-inspired Robotics and the Human Frontier Science Program (Dlife project, grant no. RGP0002/2017). We would like to thank Nienke Bijma for discussions on the functional morphology and biomechanics of dung beetles. We thank Thirawat Chutong for help with the experiments. We also thank Emily Baird and Marie Dacke for providing the video footage of dung beetles and for their insightful discussions, Peter Billeschou for his original robot hardware development, and all for their valuable research collaboration under the Dlife project.

Received: ((will be filled in by the editorial staff))

Revised: ((will be filled in by the editorial staff))

Published online: ((will be filled in by the editorial staff))

References

[1] J. Lee, J. Hwangbo, L. Wellhausen, V. Koltun, M. Hutter, Science Robotics 2020, 5986, 1.
[2] C. Yang, K. Yuan, Q. Zhu, W. Yu, Z. Li, Science Robotics 2020, 5, eabb2174.
[3] T. Miki, J. Lee, J. Hwangbo, L. Wellhausen, V. Koltun, M. Hutter, Science Robotics 2022, 7, eabk2822.
[4] J. Mu, Y. Han, G. Tan, X. Zhao, Science Advances 2015, 1, e1500533.
[5] Q. He, J. Liu, X. Sun, Y. Chen, S. Li, X. Feng, Science Advances 2023, 9, eade9247.
[6] T. Weihmann, Science Advances 2018, 4, eaat3721.
[7] Y. Gong, G. Sun, A. Nair, A. Bidwai, R. CS, J. Grezmak, G. Sartoretti, K. A. Daltorio, Front. Mech. Eng. 2023, 9, 1142421.
[8] J.-P. Sleiman, F. Farshidian, M. Hutter, Science Robotics 2023, 8, eadg5014.
[9] Boston Dynamics, available at https://www.youtube.com/watch?v=fUyU3lKzoio, accessed 2018.
[10] S. Zimmermann, R. Poranne, S. Coros, "Go Fetch! - Dynamic Grasps using Boston Dynamics Spot with External Robotic Arm", in Proc. 2021 IEEE International Conference on Robotics and Automation (ICRA), IEEE, 2021, 4488.
[11] C. D. Bellicoso, K. Krämer, M. Stäuble, D. Sako, F. Jenelten, M. Bjelonic, M. Hutter, "ALMA - articulated locomotion and manipulation for a torque-controllable robot", in Proc. 2019 International Conference on Robotics and Automation (ICRA), IEEE, 2019, 8477.
[12] Z. Fu, X. Cheng, D. Pathak, "Deep Whole-Body Control: Learning a Unified Policy for Manipulation and Locomotion", in Proc. 6th Annual Conference on Robot Learning (CoRL), 2022, 138.
[13] A. Roennau, G. Heppner, M. Nowicki, R. Dillmann, "LAURON V: A versatile six-legged walking robot with advanced maneuverability", in Proc. 2014 IEEE/ASME International Conference on Advanced Intelligent Mechatronics, IEEE, 2014, 82.

[14] M. P. Polverini, A. Laurenzi, E. M. Hoffman, F. Ruscelli, N. G. Tsagarakis, IEEE Robot. Autom. Lett. 2020, 5, 859.

[15] B. U. Rehman, M. Focchi, J. Lee, H. Dallali, D. G. Caldwell, C. Semini, "Towards a multi-legged mobile manipulator", in Proc. 2016 IEEE International Conference on Robotics and Automation (ICRA), IEEE, 2016, 3618.

[16] R. Parosi, M. Risiglione, D. G. Caldwell, C. Semini, V. Barasuol, Kinematically-decoupled impedance control for fast object visual servoing and grasping on quadruped manipulators. in Proc. 2023 IEEE/RSJ International Conference on Intelligent Robots and Systems (IROS), 2023.

[17] N. Koyachi, H. Adachi, M. Izumi, T. Hirose, "Control of walk and manipulation by a hexapod with integrated limb mechanism: MELMANTIS-1", in Proc. 2002 IEEE International Conference on Robotics and Automation, IEEE, 2002, 3553.

[18] N. Koyachi, H. Adachi, T. Arai, M. Izumi, T. Hirose, N. Senjo, R. Murata, J. Robot. Soc. Japan 2004, 22, 411.

[19] K. Inoue, K. Ooe, S. Lee, "Pushing methods for working six-legged robots capable of locomotion and manipulation in three modes", in Proc. 2010 IEEE International Conference on Robotics and Automation (ICRA), IEEE, 2010, 4742.

[20] X. Cheng, A. Kumar, D. Pathak, "Legs as Manipulator: Pushing Quadrupedal Agility Beyond Locomotion", in Proc. 2023 IEEE International Conference on Robotics and Automation (ICRA), IEEE, 2023, 5106.

[21] M. Sombolestan, Q. Nguyen, "Hierarchical adaptive loco-manipulation control for quadruped robots", in Proc. 2023 IEEE International Conference on Robotics and Automation (ICRA), IEEE, 2023, 12156.

[22] H. Deng, G. Xin, G. Zhong, M. Mistry, Object carrying of hexapod robots with integrated mechanism of leg and arm. *Robot. Comput.-Integr. Manuf.* **54**, 145–155 (2018).

[23] T. Takubo, T. Arai, K. Inoue, H. Ochi, T. Konishi, T. Tsurutani, Y. Hayashibara, E. Koyanagi, Integrated limb mechanism robot asterisk. *J. Robot. Mechatron.* **18**, 203–214 (2006).

[24] G. Takeo, T. Takubo, K. Ohara, Y. Mae, T. Arai, Internal force control for rolling operation of polygonal prism. *2009 IEEE International Conference on Robotics and Biomimetics (ROBIO), 586–591* (IEEE, 2009).

[25] G. Takeo, T. Takubo, K. Ohara, Y. Mae, T. Arai, Rotation control of polygonal prism by multi-legged robot. *2010 11th IEEE International Workshop on Advanced Motion Control (AMC), 601–606* (IEEE, 2010).

[26] J. Whitman, S. Su, S. Coros, A. Ansari, H. Choset, Generating gaits for simultaneous locomotion and manipulation. *2017 IEEE/RSJ International Conference on Intelligent Robots and Systems (IROS)*, 2723–2729 (IEEE, 2017).

[27] B. Leung, P. Billeschou, P. Manoonpong, *IEEE Trans. on Cybern.* 1–14 (2023), doi: 10.1109/TCYB.2023.3249467.

[28] M. Dacke, B. El Jundi, Y. Gagnon, A. Yilmaz, M. Byrne, E. Baird, A dung beetle that path integrates without the use of landmarks. *Animal Cogn.* **23**, 1161–1175 (2020).

[29] M. Dacke, E. Baird, B. El Jundi, E. J. Warrant, M. Byrne, How dung beetles steer straight. *Annu. Rev. Entomol.* **66**, 243–256 (2021). PMID: 32822556.

[30] B. Leung, N. Bijma, E. Baird, M. Dacke, S. Gorb, P. Manoonpong, Rules for the leg coordination of dung beetle ball rolling behaviour. *Sci. Rep.* **10**, 1–8 (2020).

[31] E. G. Matthews, Observations on the ball-rolling behavior of canthon pilularius (l.) (coleoptera, scarabaeidae). *Psyche: J. Entomol.* **70**, 75–93 (1963).

[32] B. Leung, N. Bijma, E. Baird, M. Dacke, S. Gorb, P. Manoonpong, Gait Adaptation of a Dung Beetle Rolling a Ball up a Slope. *2021 International Symposium on Adaptive Motion of Animals and Machines (AMAM),* **2**, 2–3 (2021).

[33] P. Manoonpong, H. Rajabi, J. C. Larsen, S. S. Raoufi, N. Asawalertsak, J. Homchanthanakul, H. T. Tramsen, A. Darvizeh, S. N. Gorb, *Adv. Intell. Syst.* **4**, 2100133 (2022).

[34] S. Channel, Dung beetle rolls enormous dung ball with difficulty (4k); available at https://www.youtube.com/watch?v=xNjymt6oCcQ.

[35] S. Gorb, *Attachment Devices of Insect Cuticle* (Springer Dordrecht, 2001).

[36] H. T. Tramsen, S. N. Gorb, H. Zhang, P. Manoonpong, Z. Dai, L. Heepe, Inversion of friction anisotropy in a bio-inspired asymmetrically structured surface. *J. Royal Soc. Interface* **15**, 20170629 (2018).

[37] J. Wang, C. Hu, Y. Wang, Y. Zhu, Dynamics learning with object-centric interaction networks for robot manipulation. IEEE Access 9, 68277–68288 (2021).

[38] Y. J. Ma, W. Liang, H. Wang, S. Wang, Y. Zhu, L. Fan, O. Bastani, D. Jayaraman, in Proc. Robotics: Science and Systems (RSS), 2024.

[39] F. Shi, T. Homberger, J. Lee, T. Miki, M. Zhao, F. Farshidian, K. Okada, M. Inaba, M. Hutter, Circus ANYmal: A quadruped learning dexterous manipulation with its limbs. in Proc. 2021 IEEE International Conference on Robotics and Automation (ICRA), 2316–2323 (IEEE, 2021).

[40] Y. Zhou, M. Liu, C. Song, J. Luo, Kirin: A Quadruped Robot with High Payload Carrying Capability. available at https://arxiv.org/abs/2202.08620, accessed July 2023 (2022).

[41] N. A. M. Khairussaleh, R. Omar, S. M. Aris, M. F. M. Nor, M. A. M. Saidi, N. M. A. N. M. Mahari, in Proc. IOP Conference Series: Earth and Environmental Science, IOP Publishing, 2023, 1140, 012016.

[42] L. A. Smith, W. B. Anthony, E. S. Renoll, J. L. Stallings, Hay in Round and Conventional Bale Systems, Auburn: Agricultural Experiment Station, Auburn University, 1975.

[43] P. Billeschou, N. N. Bijma, L. B. Larsen, S. N. Gorb, J. C. Larsen, P. Manoonpong, Appl. Sci. 2020, 10, 6986.

[44] A. Spröwitz, A. Tuleu, M. Vespignani, M. Ajallooeian, E. Badri, A. J. Ijspeert, Int. J. Rob. Res. 2013, 32, 932.

[45] C. Ji, F. Li, X. Chao, Free gait transition and stable motion generation using CPG-based locomotion control for hexapod robots. (2024).

[46] M. Sombolestan, Q. Nguyen, Hierarchical adaptive loco-manipulation control for quadruped robots. in Proc. 2023 IEEE International Conference on Robotics and Automation (ICRA), 12156–12162 (2023).

[47] Spot | Boston Dynamics, Boston Dynamics, https://bostondynamics.com/products/spot/, accessed: September 4, 2024.

[48] C. Lin, X. Liu, Y. Yang, Y. Niu, W. Yu, T. Zhang, J. Tan, B. Boots, D. Zhao, LocoMan: Advancing Versatile Quadrupedal Dexterity with Lightweight Loco-Manipulators. arXiv Preprint arXiv:2403.18197 (2024).

[49] H. Ferrolho, W. Merkt, V. Ivan, W. Wolfslag, S. Vijayakumar, Optimizing dynamic trajectories for robustness to disturbances using polytopic projections. in Proc. 2020 IEEE/RSJ International Conference on Intelligent Robots and Systems (IROS), 7477–7484 (IEEE, 2020).

[50] H. Ferrolho, V. Ivan, W. Merkt, I. Havoutis, S. Vijayakumar, Roloma: Robust loco-manipulation for quadruped robots with arms. Autonomous Robots 47, 1463–1481 (2023).